%% file: main.tex
\documentclass[journal, 10pt, twoside]{IEEEtran}
\usepackage{cite}
\usepackage{makecell}
\usepackage{booktabs}
\usepackage{multirow}
\usepackage{threeparttable}
\usepackage[switch]{lineno}
\usepackage{multicol}
\input{packages.tex}
\newcommand\clr[1]{{\color{black}{#1}}}
\newcommand\cl[1]{{\color{black}{#1}}}

\newcolumntype{L}[1]{>{\raggedright\arraybackslash}p{#1}}
\newcolumntype{M}[1]{>{\centering\arraybackslash}m{#1}}

\newcommand{\settablefont}{\fontsize{5.5}{8.6}\selectfont}
\begin{document}

\title{UDTIRI: An Online Open-Source\\Intelligent Road Inspection Benchmark Suite}
\author{Sicen Guo$^{\orcidicon{0009-0000-8079-8056}\,}$, Jiahang Li$^{\orcidicon{0009-0005-8379-249X}\,}$, Yi Feng$^{\orcidicon{0009-0005-4885-0850}\,}$, Dacheng Zhou, Denghuang Zhang, Chen Chen, Shuai Su$^{\orcidicon{0000-0001-6144-8923}\,}$, \\Xingyi Zhu$^{\orcidicon{0000-0002-0822-6261}\,}$, Qijun Chen$^{\orcidicon{0000-0001-5644-1188}\,}$,~\IEEEmembership{Senior Member,~IEEE}, Rui Fan$^{\orcidicon{0000-0003-2593-6596}\,}$,~\IEEEmembership{{Senior} Member,~IEEE}

\thanks{
This research was supported in part by the National Natural Science Foundation of China under Grants 62233013 and 52278455, the Science and Technology Commission of Shanghai Municipal under Grant 22511104500, the Shuguang Program of Shanghai Education Development Foundation and Shanghai Municipal Education Commission under Grant 21SG24, the International Cooperation Project of Science and Technology Commission of Shanghai Municipality under Grant 22210710700, and the Fundamental Research Funds for the Central Universities. \textit{(Sicen Guo and Jiahang Li are joint first authors.) ({Corresponding author: Rui Fan.})} }

\thanks{
Sicen Guo, Jiahang Li, Yi Feng, and Rui Fan are with the Machine Intelligence \& Autonomous Systems (MIAS) Group, the College of Electronics \& Information Engineering, Shanghai Research Institute for Intelligent Autonomous Systems, the State Key Laboratory of Intelligent Autonomous Systems, and Frontiers Science Center for Intelligent Autonomous Systems, Tongji University, Shanghai 201804, China (e-mail: \{guosicen, lijiahang617, fengyi0109, rfan\}@tongji.edu.cn)
}

\thanks{
Dacheng Zhou, Chen Chen, and Denghuang Zhang are with the Department of Control Science \& Engineering, Tongji University, Shanghai 201804, China (e-mail: zhoudacheng20@tongji.edu.cn; \{ccsama0109, zhangdenghuang666\}@gmail.com) 
}

\thanks{
Shuai Su and Qijun Chen are with the Robotics \& Artificial Intelligence Laboratory (RAIL), the College of Electronics \& Information Engineering, Tongji University, Shanghai 201804, China (e-mail: \{sushuai, qjchen\}@tongji.edu.cn)
}

\thanks{Xingyi Zhu is with the Department of Road \& Airport Engineering and the Key Laboratory of Road \& Traffic Engineering of Ministry of Education, Tongji University, Shanghai 200092, China. (e-mail: { zhuxingyi66@tongji.edu.cn})}

\thanks{
The UDTIRI online benchmark suite is accessible at \url{https://udtiri.com}.
}

\thanks{
Color versions of one or more figures in this article are available at
https://doi.org/xx.xxxx/TITS.20xx.xxxxxxx.
}

\thanks{
Digital Object Identifier xx.xxxx/TITS.20xx.xxxxxxx
}

}

\markboth{IEEE Transactions on Intelligent Transportation Systems}
	{Guo \MakeLowercase{\textit{et al.}}: UDTIRI: An Online Open-Source Intelligent Road Inspection Benchmark Suite}
\maketitle

\begin{abstract}
In the nascent domain of urban digital twins (UDT), the prospects for leveraging cutting-edge deep learning techniques are vast and compelling. Particularly within the specialized area of intelligent road inspection (IRI), a noticeable gap exists, underscored by the current dearth of dedicated research efforts and the lack of large-scale well-annotated datasets. To foster advancements in this burgeoning field, we have launched an online open-source benchmark suite, referred to as UDTIRI. Along with this article, we introduce the road pothole detection task, the first online competition published within this benchmark suite. This task provides a well-annotated dataset, comprising 1,000 RGB images and their pixel/instance-level ground-truth annotations, captured in diverse real-world scenarios under different illumination and weather conditions. Our benchmark provides a systematic and thorough evaluation of state-of-the-art object detection, semantic segmentation, and instance segmentation networks, developed based on either convolutional neural networks or Transformers. We anticipate that our benchmark will serve as a catalyst for the integration of advanced UDT techniques into IRI. By providing algorithms with a more comprehensive understanding of diverse road conditions, we seek to unlock their untapped potential and foster innovation in this critical domain.
\end{abstract}

\begin{IEEEkeywords}
Urban digital twins, deep learning, intelligent road inspection, benchmark suite, road pothole detection.
\end{IEEEkeywords}

\section{Introduction}
\label{sec.introduction}

\IEEEPARstart{D}{igital} twin (DT) represents the forefront of technology, including innovative algorithms that bridge physical systems with digital networks \cite{cheng2018cyber}. This integration blurs the boundaries between the physical and digital domains, heralding a new era of interconnectedness and real-time analytics \cite{guo2023digital}. With the rapid pace of digital transformation, the applications of DT technology expand across diverse sectors, from smart manufacturing \cite{son2022past} to intelligent urban planning \cite{xia2022study} and advanced medical healthcare \cite{alazab2022digital}. An urban digital twin (UDT) is fashioned by encoding the semantic and geospatial properties of urban entities, such as buildings and roads \cite{lei2023challenges}. These digital replicas of physical urban infrastructures are indispensable to fulfill a diverse range of needs and uses, as exemplified by applications such as intelligent road inspection (IRI) \cite{fan2022urban}.

\begin{figure}[t!]
		\centering
		\includegraphics[width=0.49\textwidth]{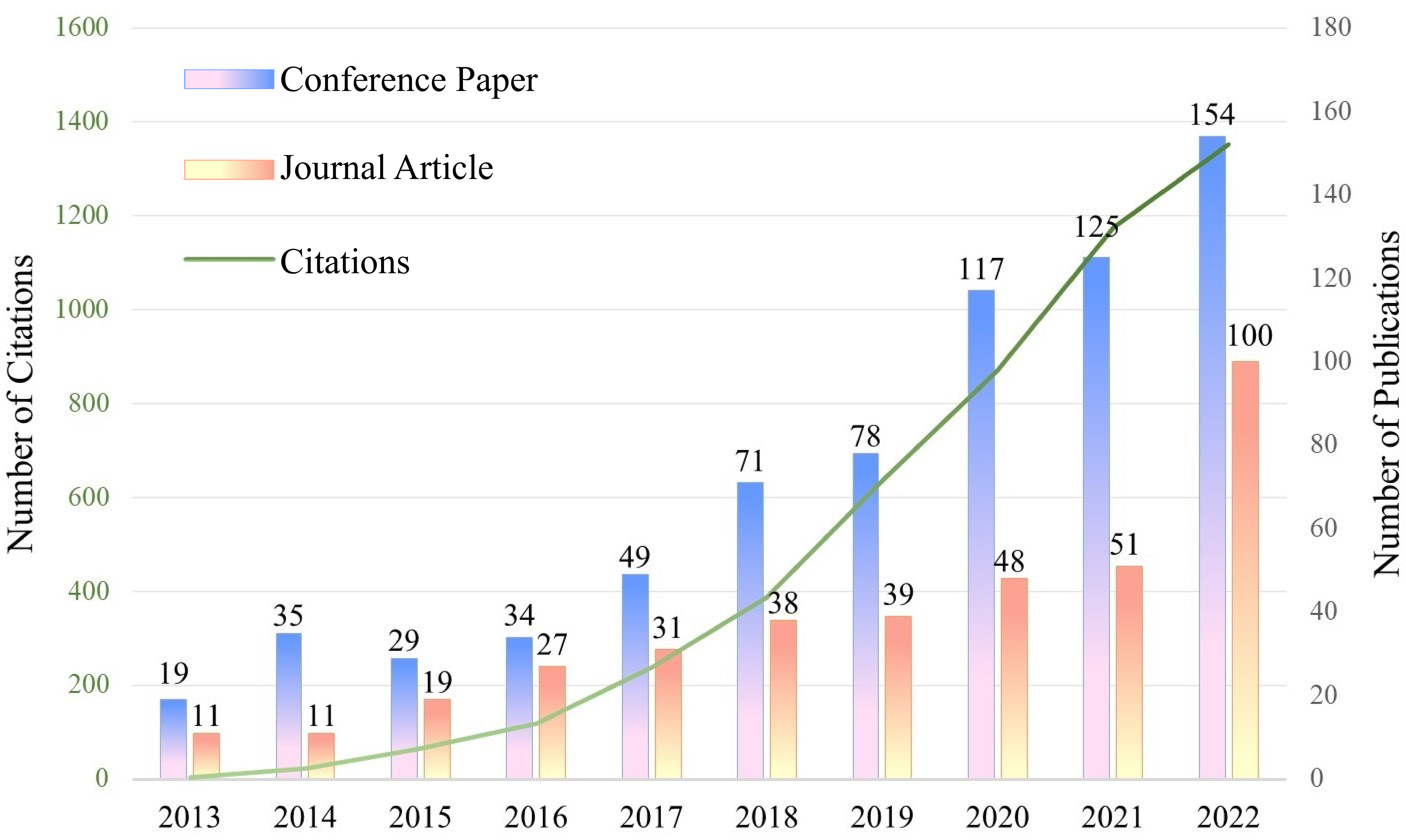}
		\centering
		\caption{Publication and citation trends for road pothole detection over the past decade. Conference papers are sourced from the Engineering Village database (webpage: \url{engineeringvillage.com/}), while journal articles and citations are sourced from the Web of Science database (webpage: \url{webofscience.com}).}
		\label{fig.publish}
\end{figure} 

\begin{figure*}[t!]
		\centering
		\includegraphics[width=1\textwidth]{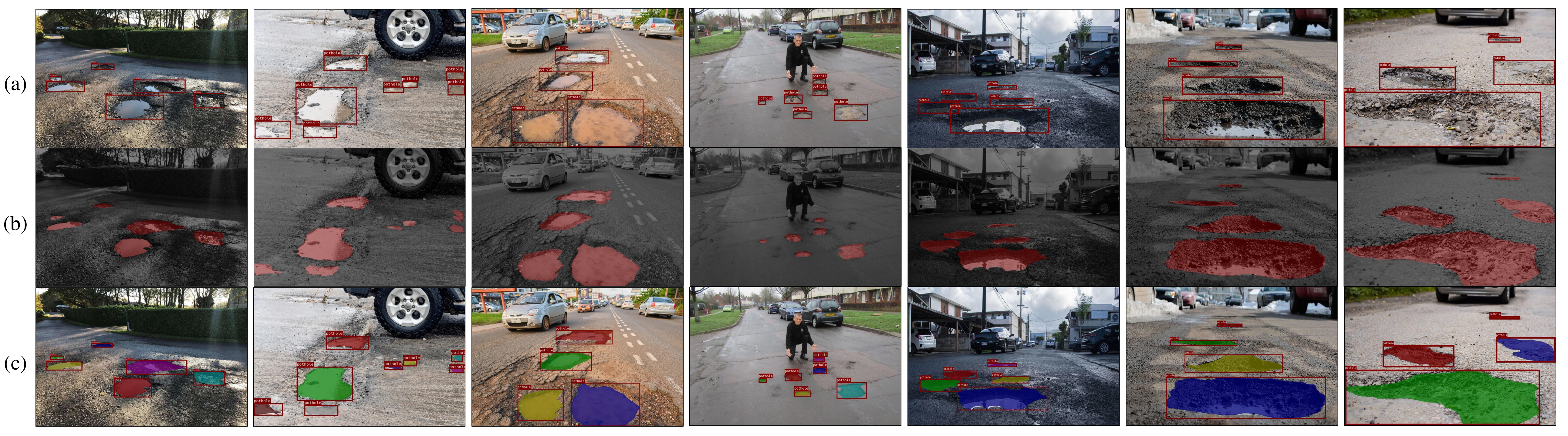}
		\centering
		\caption{Examples of the ground-truth annotations for the road pothole detection competition within the UDTIRI benchmark suite: (a) object detection; (b) semantic segmentation; (c) instance segmentation.}
		\label{fig.datasetpic}
\end{figure*}

Traditional road inspection is typically conducted by structural engineers or certified inspectors \cite{fan2023autonomous}. However, this process is fraught with challenges: it is perilous, inefficient, costly, and tedious \cite{li2023roadformer}. Additionally, the road inspection results are often qualitative and subjective, relying solely on the expertise of the individual inspectors \cite{mathavan2015review}. With the advancement of UDT techniques, especially deep neural networks, there is an increasing appetite for data-driven IRI systems, which generally undertake two primary tasks \cite{ma2022computer}: (1) road data acquisition \cite{fan2018road} and (2) road damage detection \cite{fan2019pothole}. Developing a comprehensive, open-source, and well-annotated online benchmark suite for evaluating UDT techniques applied to IRI is, therefore, of paramount significance to the intelligent transportation society.

Potholes, among the most prevalent types of road damage, are considerably large structural defects on the road surface \cite{fan2021rethinking}. Detecting these defects is not only vital for proactive urban road maintenance but also imperative for autonomous driving \cite{fan2020we}. However, current autonomous driving perception systems prioritize the detection of large objects of interest, \textit{e.g.}, pedestrians, traffic signs, and vehicles, often sidelining road damage. Nevertheless, driving quality, vehicle maneuverability, fuel consumption, and tire longevity are all related to road conditions. Therefore, accurate and efficient detection of road potholes is crucial for improving both driving comfort and safety \cite{fan2022urban}.

A thorough search of the relevant literature results in Fig. \ref{fig.publish}, which highlights the growing interest in road pothole detection over the past decade, affirming its position as a burgeoning research topic. Our recent survey article \cite{ma2022computer} categorizes the existing road pothole detection algorithms into three groups: (1) classical 2-D image processing-based, (2) 3-D point cloud modeling and segmentation-based, and (3) data-driven approaches. The first category of algorithms generally utilizes explicit image processing algorithms to segment road RGB or disparity/depth images \cite{koch2015review}. Such algorithms are often computationally demanding and sensitive to various environmental factors, notably illumination and weather conditions \cite{jahanshahi2013unsupervised}. Additionally, the irregular shapes of road potholes render the geometric assumptions made in such approaches occasionally infeasible. Therefore, 3-D point cloud modeling and segmentation-based algorithms have become popular choices for road pothole detection \cite{zhang2013advanced}. These algorithms typically consider the 3-D road point clouds, captured using a range sensor, as a quadratic surface. The raw point clouds are then segmented by comparing the differences between the modeled surface and the actual data \cite{fan2019road}. However, these algorithms are still relatively underutilized. This is primarily because accurate 3-D road imaging is costly, and real-world road surfaces can be highly irregular and uneven, sometimes rendering these techniques impractical. Data-driven approaches, typically developed based on convolutional neural networks (CNNs), have emerged as frontrunners, delivering compelling road pothole detection results \cite{fan2020we, dhiman2019pothole, li2023roadformer, fan2021graph}. 

Over the past decade, the advent of several online benchmark suites, such as KITTI \cite{geiger2012we} and Cityscapes \cite{cordts2016cityscapes}, has been playing a pivotal role in advancing the performance of general visual perception algorithms. However, despite the abundance of such datasets for general computer vision research, the specific domain of IRI, especially when underpinned by cutting-edge UDT techniques, remains relatively underexplored. A primary reason is that road defects such as potholes are not ubiquitous, making the creation of large-scale datasets inherently challenging \cite{fan2021rethinking}. Furthermore, most existing approaches in this area simply apply transfer learning to fine-tune state-of-the-art (SoTA) object detection or semantic segmentation models on relatively small road inspection datasets. Moreover, previous studies in this research area typically reported results based on experiments where datasets were split randomly. Comparing algorithms on the same dataset without consistent data splits can skew results, as performance may be influenced by overlapping training and validation data distributions. Finally, while all existing road pothole detection datasets are created for either instance-level object detection or pixel-level semantic segmentation, there is a noticeable absence of large-scale datasets designed to accommodate both tasks simultaneously via instance segmentation. Therefore, developing an online open-source benchmark suite comprising a variety of IRI tasks, including but not limited to road surface 3D reconstruction and road damage detection, is a popular area of research that requires more attention.

In this article, we introduce the \uline{\textbf{U}rban \textbf{D}igital \textbf{T}wins for \textbf{I}ntelligent \textbf{R}oad \textbf{I}nspection (\textbf{UDTIRI})} online benchmark suite, accessible at \url{https://udtiri.com/}. Road pothole detection, the first online competition launched within this benchmark suite, provides researchers with a large-scale, well-annotated dataset for comprehensive evaluation of object detection, semantic segmentation, and instance segmentation networks, designed for this specific task. Similar to KITTI \cite{geiger2012we} and Cityscapes \cite{cordts2016cityscapes}, the ground-truth annotations (see Fig. \ref{fig.datasetpic}) are available for model training and validation, while the evaluation metrics on the test set can be acquired by uploading results to the UDTIRI benchmark suite. To set a reference point, we have conducted extensive experiments with 14 SoTA object detection networks, 30 SoTA semantic segmentation networks, and ten SoTA instance segmentation networks, providing baseline results for road pothole detection. With additional online competitions launched within this benchmark suite in the near future, we believe that it will serve as a catalyst for the integration of cutting-edge UDT methodologies into IRI.

The remainder of this article is structured as follows: Sect. \ref{sec.literature_review} presents a comprehensive review of SoTA object detection, semantic segmentation, and instance segmentation networks. Sect. \ref{sec.udtiri_benchmark} details our UDTIRI benchmark suite and road pothole detection dataset. Sect. \ref{sec.experiments} presents the conducted experiments and provides both qualitative and quantitative comparisons of the SoTA networks. Sect. \ref{sec.discussion} discusses the performance and potential limitations of the compared networks. Finally, we summarize our contributions in Sect. \ref{sec.Conclusion}.

\section{Literature Review}
\label{sec.literature_review}

\subsection{Objection Detection Networks} 
\label{sec.object_detection}

\clr{Existing CNN-based object detection methods are primarily divided into two categories: two-stage and one-stage ones. Two-stage methods first generate regions of interest (RoIs), which are then refined for both object classification and bounding box regression. Nevertheless, the sequential nature of two-stage methods can result in slower inference speeds. On the other hand, one-stage methods formulate object detection as a direct bounding box regression task, typically achieving higher computational efficiency yet sometimes sacrificing detection accuracy compared to two-stage methods.

As a pioneering two-stage approach, R-CNN \cite{girshick2014rich} introduces region proposals, leverages a CNN to extract features from these proposals, and classifies these proposals with a support vector machine (SVM). Unfortunately, its multi-stage training pipeline places an exceptionally heavy computational burden. To address this limitation, Fast R-CNN \cite{Girshick_2015_ICCV} streamlines the object detection process by pooling CNN features associated with each region proposal, significantly improving overall efficiency by sharing computations for overlapping regions. However, it still relies on the relatively slow selective search for region proposal generation. Therefore, Faster R-CNN \cite{ren2015faster} addresses this drawback by introducing a region proposal network (RPN), which directly generates region proposals, thereby enabling end-to-end training.

One-stage methods, exemplified by the you only look once (YOLO) series, single-shot multi-box detector (SSD) \cite{liu2016ssd}, CenterNet \cite{zhou2019objects}, RetinaNet \cite{lin2017focal}, and EfficientDet \cite{DBLP:conf/cvpr/TanPL20}, have gained increasing attention owing to their remarkable real-time performance. These approaches can be broadly categorized as anchor-free or anchor-based. The fundamental difference between these two categories lies in their reliance on pre-defined anchors to assist object detection. Anchor-based object detection frameworks require careful anchor design to adequately capture the scale and aspect ratio of specific object classes. In contrast, anchor-free approaches forgo the use of pre-defined anchors and instead directly predict two points (top-left and bottom-right) for each object.

As the first anchor-free approach, YOLOv1 \cite{redmon2016you} simplifies object detection into a direct bounding box regression task. It utilizes a CNN for feature extraction and a fully connected layer to regress object bounding box coordinates and classes. Furthermore, YOLOv6 \cite{li2022yolov6} 
incorporates the SCYLLA-IoU (SIoU) \cite{gevorgyan2022siou} bounding box regression loss, which leads to improved object detection accuracy when compared to earlier anchor-based models, such as YOLOv2 through YOLOv5. Additionally, CenterNet \cite{zhou2019objects} represents each object as a single point at the bounding box center, instead of regressing the entire bounding box. This point-based representation generally enhances model generalizability and can be readily extended to related tasks, such as 3D object detection, instance segmentation, and keypoint estimation. 

Among the anchor-based YOLO series, YOLOv2 \cite{redmon2017yolo9000} employs a high-resolution classifier and anchor boxes to improve object detection accuracy. Nevertheless, YOLOv2 struggles with fine-grained localizations, overlapping objects, and potential loss of spatial information, even though it achieves real-time performance. YOLOv3 \cite{redmon2018yolov3} addresses these limitations by making predictions across three different scales, capturing more comprehensive semantics to improve object detection performance. YOLOv4 \cite{bochkovskiy2020yolov4} incorporates a cross-stage partial network \cite{wang2020cspnet} into its backbone, notably enhancing learning capability while also reducing computational complexity compared to YOLOv2 and YOLOv3. Moreover, YOLOv5 \cite{jocher2022ultralytics} utilizes an embedded anchor box selection mechanism to improve training and inference speed compared to YOLOv4. YOLOv7 \cite{wang2022yolov7} introduces an extended efficient layer aggregation network to further boost inference speed. 
It achieves the most favorable trade-off between efficiency and accuracy when compared to all previous versions of YOLO. SSD \cite{liu2016ssd} was also developed with the primary aim of achieving a balance between speed and accuracy by utilizing pre-defined anchors and multi-scale features. Additionally, EfficientDet \cite{DBLP:conf/cvpr/TanPL20} optimizes the trade-off among model complexity, speed, and accuracy by adjusting network depth, width, and input resolution. Unfortunately, most one-stage detectors still lag behind two-stage models in accuracy, primarily due to sensitivity to foreground-background class imbalances.
RetinaNet \cite{lin2017focal} addresses this limitation by employing focal loss to focus on ``hard'' samples,  allowing it to maintain high-speed processing while remaining competitive with SoTA one-stage methods.

DETR \cite{carion2020end} employs an encoder-decoder Transformer architecture alongwith a set-based global loss to produce an optimal bipartite matching between predicted and ground-truth objects. This loss function uniquely associates each prediction with a specific target object, ensuring invariance to the order of predictions. To overcome the challenges of slow training convergence and high computational complexity inherent in DETR, deformable DETR \cite{zhu2020deformable} incorporates deformable convolutions for sparse spatial sampling. Furthermore, it utilizes a deformable attention module to concentrate on a small set of key sampling points around a reference point, regardless of the spatial size of the feature maps. This addresses a limitation in the standard Transformer attention mechanism, which typically considers all possible spatial locations.
}

\subsection{Semantic Segmentation Networks}

\clr{Fully convolutional network (FCN) \cite{long2015fully} marked a pioneering milestone in the use of CNNs for end-to-end semantic segmentation. However, its segmentation does not fully account for pixel relationships, resulting in segmentation results that lack spatial consistency. Additionally, FCN also significantly amplifies memory usage and computational complexity. To address these limitations, Fast FCN \cite{Wu2019} extracts high-resolution feature maps through upsampling convolutions. This approach effectively addresses the spatial inconsistency issue and significantly reduces computational  complexity by more than threefold.

Recent approaches \cite{kirillov2019panoptic, Xiao2018, florian2017rethinking, Chen2018, He2019a, Zhao2017} have made significant strides in enhancing performance by expanding the receptive fields using pyramid-based multi-resolution techniques. Pyramid scene parsing network (PSPNet) \cite{Zhao2017} performs spatial pyramid pooling (SPP) at multiple scales, achieving exceptional performance across several semantic segmentation benchmarks. Similarly, based on Mask R-CNN \cite{he2017mask} and feature pyramid network (FPN) \cite{lin2017feature}, panoptic FPN \cite{kirillov2019panoptic} utilizes a lightweight semantic segmentation branch for dense pixel prediction. Furthermore, DeepLabv3 \cite{florian2017rethinking} employs several parallel atrous SPP (ASPP) modules to gather contextual information across multiple scales. Nevertheless, the stride operations used in DeepLabv3 may lead to the loss of object boundary details. To address this limitation, DeepLabv3+ \cite{Chen2018} introduces a concise yet effective decoder into DeepLabv3, significantly improving semantic segmentation results, particularly along label boundaries. Additionally, dynamic multi-scale network (DMNet) \cite{He2019a} learns variable-scale features through dynamic multi-scale filters. It is more adaptable and flexible, as each branch can capture a unique scale of features relevant to the input image.

U-Net \cite{ronneberger2015u} features a U-shaped encoder-decoder structure, originally designed for biomedical image segmentation problems. In contrast to symmetric encoder-encoder architectures utilized in the following studies \cite{ronneberger2015u, Chaurasia2017, Badrinarayanan2017}, efficient neural network (ENet) \cite{Paszke2016} adopts a larger encoder paired with a smaller decoder. The encoder effectively handles data with lower resolutions, thereby providing the decoder with fine-grained features. Subsequently, the decoder samples these features and refines boundary details. SegResNet \cite{Myronenko2018}, another asymmetric encoder-encoder architecture, replaces the encoder of SegNet \cite{Badrinarayanan2017} with ResNet blocks. Moreover, it incorporates a variational autoencoder branch to regularize the shared encoder by reconstructing input images. Unlike these prior arts \cite{ronneberger2015u,long2015fully,Chaurasia2017,Badrinarayanan2017,Paszke2016,Myronenko2018} that focus on the recovery of high-resolution feature maps from low-resolution representations, high-resolution network (HRNet) \cite{Sun2019} maintains high-resolution representations throughout the entire feature extraction and fusion process, resulting in more accurate predictions, achieved through progressive and repetitive multi-scale feature fusion, conducted by multi-resolution sub-networks in parallel.

Attention mechanisms have been playing a pivotal role in recent semantic segmentation networks \cite{jia2023tfgnet, jia2023enhancing}. As two notable approaches, the non-local neural network (Non-local) \cite{Wang2018} obtains the attention mask by computing the correlation matrix between each point in the feature maps, and PSANet \cite{Zhao2018} learns the mask by aggregating context information for each specific point in a self-adaptive manner. However, the extensive computational demands of attention mechanisms have limited their application in various real-world scenarios. To overcome this challenge, the asymmetric non-local neural network (ANN) \cite{zhu2019asymmetric} samples only a few representative points from the feature maps, significantly reducing computational complexity. Additionally, the attention computation can be decomposed into a pair-wise term and a unary term, which can be challenging to learn independently. The disentangled non-local network (DNLNet) \cite{Yin2020} addresses this issue by decoupling the tight relationship between these two components.
Most attention-based approaches \cite{Zhao2018,Fu2019} use adaptive weights to compute pair-wise similarity or learn pixel-wise attention maps. However, they tend to overlook the importance of global guidance from the feature extractors. To address this limitation, adaptive pyramid context network (APCNet) \cite{He2019} estimates the degree of sub-region contribution from local and global representations and leverages multi-scale representations with a feature pyramid, resulting in improved overall performance.

While attention mechanisms have demonstrated superior performance compared to ASPP \cite{florian2017rethinking}, large convolutional kernels, and stacked convolutional layers, their heightened demand for GPU memory can often be prohibitively expensive. Therefore, several networks have emerged with a primary focus on further minimizing these computational requirements. Criss-cross network (CCNet) \cite{Huang2019} introduces fully spatial attention, while interlaced sparse self-attention network (ISANet) \cite{huang2000interlaced} factorizes the dense affinity matrix into the product of two sparse affinity matrices. Furthermore, instead of treating all pixels as reconstruction bases \cite{Wang2018, Zhao2018}, the expectation maximization attention network (EMANet) \cite{Li2019} finds a more compact basis set, leading to a substantial reduction in computational complexity. Additionally, context encoding network (ENCNet) \cite{Zhang2018} selectively highlights the class-dependent feature maps, thereby infusing the scene-relevant prior information into the network. This technique simplifies the generation of large attention maps while notably reducing memory consumption.

Vision Transformer (ViT) \cite{dosovitskiy2020image} has been gaining momentum in recent years. Swin Transformer \cite{Liu2021}, Segmenter \cite{Strudel2021}, and Twins \cite{Chu2021} are all developed based on ViT \cite{dosovitskiy2020image}. Swin Transformer designs a hierarchical Transformer architecture that computes representations with shifted windows. Inspired by DETR \cite{NicolasCarion2020}, Segmenter develops a mask Transformer decoder, capable of capturing global context at each layer during both encoding and decoding stages. To improve semantic segmentation at both global and local scales, Twins \cite{Chu2021} adopts a two-branch architecture: one captures global contextual information, while the other one focuses on the local boundary details of the segmented regions. Furthermore, SegFormer \cite{Xie2021} aggregates information from various layers, effectively combining both local and global attention to produce robust and powerful representations. To improve the learned representations, ResNeSt \cite{Zhang2022} combines channel-wise attention with multi-path representation into a single unified split-attention block. Similar to the self-attention mechanism used in ViT, object-contextual representation (OCR) \cite{yuan2020object} characterizes pixels by exploiting the representations of corresponding object classes. The conventional multi-scale context schemes, such as SPP \cite{Lazebnik2006} and ASPP \cite{florian2017rethinking}, only differentiate pixels with different spatial positions, while OCR \cite{yuan2020object} distinguishes between contextual pixels of the same object class and those of different object classes.}
   
\subsection{Instance Segmentation Networks}

\clr{
Similar to object detection networks, instance segmentation networks can also be broadly categorized as either two-stage and one-stage ones \cite{tian2022review}. The former networks \cite{he2017mask,cai2019cascade,huang2019mask,bolya2019yolact,tian2021boxinst} first detect bounding boxes for each instance and then perform pixel classification within each bounding box to generate the final mask. In contrast, one-stage networks \cite{wang2020solo,wang2020solov2,cao2019gcnet,al2022overview} directly propose prediction boxes from the input images without a region proposal step \cite{jiao2019survey}. Although one-stage methods are more suitable for applications requiring real-time performance due to their concise and efficient architectures, two-stage methods typically achieve higher segmentation accuracy, primarily attributed to the second refined stage \cite{gu2022review}.

Mask R-CNN \cite{he2017mask} is a pioneering two-stage framework that extends Faster R-CNN \cite{ren2015faster} by adding an additional branch to predict pixel-level masks in parallel with the existing branch that recognize bounding boxes. Its architecture consists of a CNN backbone, a RPN, and two heads separately for object classification and prediction. The introduction of the RoI align (abbreviated as RoIAlign) module ensures that the extracted features are correctly aligned, thus eliminating the misalignment issues caused by quantization errors present in previous methods \cite{ren2015faster}. Cascade R-CNN \cite{cai2019cascade} is another extension of Faster R-CNN \cite{ren2015faster}, which improves instance detection accuracy through a cascade of multiple stages. Unlike the methods \cite{he2017mask,ren2015faster,cai2019cascade} that usually predict mask quality score based on the confidence of instance segmentation networks, Mask Scoring R-CNN \cite{huang2019mask} takes both the instance feature and the corresponding predicted mask into account to regress the intersection over union (IoU) score for masks. This approach considers the accuracy of both semantic categories and the instance masks, presenting a novel method for scoring the instance segmentation hypotheses and offering a new perspective on the evaluation of instance segmentation performance.

Unlike Mask R-CNN \cite{he2017mask} that relies on RoI operations (typically RoIAlign) to obtain the final instance masks, YOLACT \cite{bolya2019yolact} decouples RoI detection from the feature maps used for mask prediction. Additionally, instead of using instance-wise RoIs as inputs to a network with fixed weights, CondInst \cite{tian2020conditional} employs dynamic instance-aware networks conditioned on instances. This approach offers two notable advantages: (1) it eliminates the need for RoI cropping and feature alignment through an FCN module; (2) with the increased capacity of dynamically generated conditional convolutions, the compact mask head leads to significantly faster inference speed. The same research group also proposed BoxInst \cite{tian2021boxinst}, which realizes instance segmentation with the use of two losses: (1) a surrogate loss that focuses on minimizing the discrepancy between the projections of the ground-truth box and the predicted mask, and (2) a pair-wise loss that supervises the label consistency in proximal pixels, determining whether two pixels have the same labels or not.

As a representative one-stage instance segmentation network, segmenting objects by locations (SOLO) \cite{wang2020solo} uses a similar paradigm to semantic segmentation for the instance segmentation task. Basically, the mask branch predicts soft masks for all potential objects, while the category branch subsequently determines the object classes, enabling efficient instance segmentation without utilizing RoI operations. Nonetheless, SOLO struggles to segment small instances. To overcome this limitation, SOLOv2 \cite{wang2020solov2} dynamically predicts mask kernels based on the input and assigns appropriate location categories to different pixels. Additionally, to prevent duplicate predictions, it employs ``matrix non-maximum suppression (NMS)'', which dramatically boost the model's inference speed.

Capturing contextual information and long-range dependencies is crucial for instance segmentation. Global context network (GCNet) \cite{cao2019gcnet} simplifies the non-local network \cite{wang2018non} by explicitly utilizing a query-independent attention map applicable to all query positions.
GCNet has proven to deliver impressive performance, primarily due to its capacity to model pixel-level long-range dependencies while simultaneously mapping channel-wise attention. Additionally, to address the challenge of capturing long-range dependencies, deformable convolutional network (DCN) \cite{dai2017deformable} introduces deformable convolution, which offers dynamic and learnable receptive fields, effectively adapting to the image content. It solves the inherent limitations of geometric transformations in CNNs and has demonstrated exceptional performance across various computer vision tasks.
}

\section{UDTIRI Benchmark Suite}
\label{sec.udtiri_benchmark}

\begin{figure*}[t!]
	\begin{center}
		\centering
		\includegraphics[width= 0.99\textwidth]{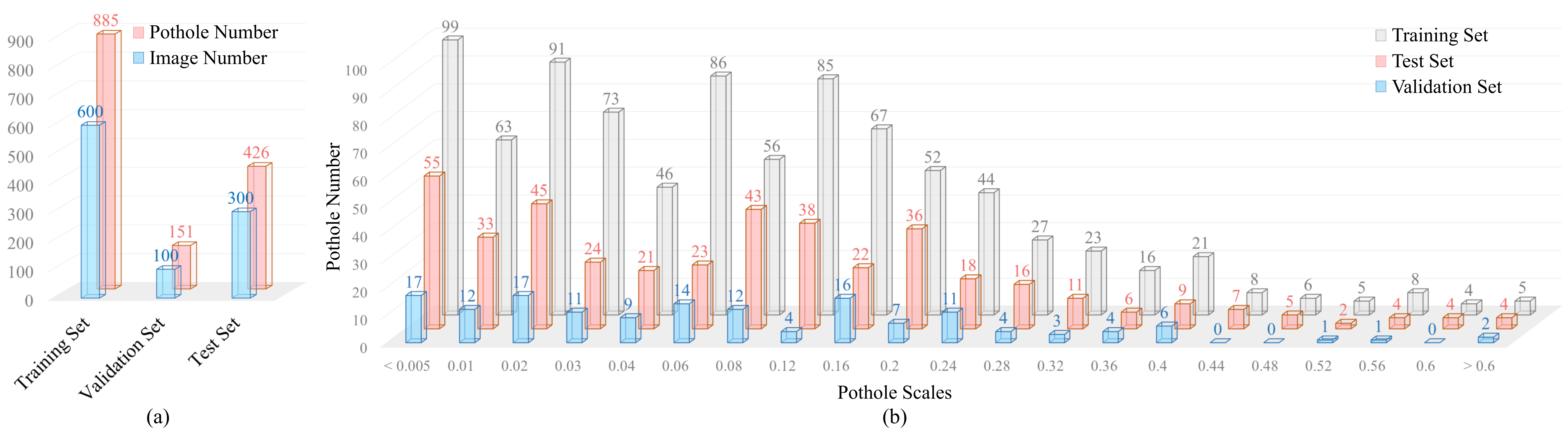}
		\centering
		\caption{Dataset characteristics: (a) a histogram showing the distribution of pothole numbers; (b) a histogram showing the distribution of pothole scales.} 
		\label{fig.trainvaltest}
	\end{center}
\end{figure*}

Our initial aim centered around the establishment of an online open-source benchmark suite, developed to provide comprehensive evaluations of cutting-edge UDT techniques applied to address IRI problems. These evaluations involve various general visual perception algorithms, including but not limited to object detection (instance-level perception), semantic segmentation (pixel-level perception), and instance segmentation (perception at both instance and pixel levels) networks. However, it is worth noting that these general models have rarely been evaluated specifically for IRI tasks, primarily due to the lack of a public dataset that includes diverse forms of ground-truth annotations and provides reasonable data partitions. Therefore in this paper, we take the initial step by launching a road pothole detection competition based on a large-scale, well-annotated, multi-purpose, real-world dataset.

Prior to introducing our newly developed road pothole detection dataset, we first provide a brief overview of the relevant existing datasets that have been created for the evaluation of visual perception algorithms. Labeling object detection ground truth is relatively inexpensive and can result in larger datasets. For example, a dataset\footnote{\url{kaggle.com/sovitrath/road-pothole-images-for-pothole-detection}} \cite{rath} was created for object detection-based pothole detection, which contains 3,777 RGB images for training and 628 RGB images for testing. However,  in the context of road inspection, our primary focus is on acquiring accurate information (\textit{e.g.}, shapes and sizes) of road potholes. This objective inherently requires accurate pixel-level annotations. In our previous work \cite{fan2019pothole}, we published the first pixel-level road pothole detection dataset\footnote{\url{github.com/ruirangerfan/stereo_pothole_datasets}}, which contains 67 collections of RGB images (resolution: 800$\times$1,312 pixels), subpixel disparity images, transformed disparity images, and ground-truth annotations. Furthermore, we published a relatively larger dataset, referred to as the Pothole-600 dataset\footnote{\url{sites.google.com/view/pothole-600}} \cite{fan2020we}, which contains 600 pairs of RGB images and transformed disparity images. While the datasets mentioned above are suitable for the evaluation of semantic segmentation algorithms, they were created under rather limited illumination and weather conditions. Moreover, the road potholes in these datasets are comparatively easy to recognize from such scenarios.

To address the absence of large-scale, multi-functional, well-annotated road pothole datasets, we collect road data with respect to diverse pothole depths, sizes, and shapes, captured using different cameras mounted on different vehicles and under various weather and illumination conditions. These comprehensive data are not only suitable for model training in each of these individual tasks but can also be leveraged for multi-task learning, thereby setting our dataset apart from existing options. Within our dataset, the road potholes have a wide range of scales, as depicted in Fig. \ref{fig.trainvaltest}. Based on these statistical analyses, we categorize the road potholes into large ($L$), medium ($M$), and small ($S$) ones. In our experiments, we conduct a comprehensive performance evaluation of object detection and instance segmentation networks with respect to the different scales of road potholes.

All the images in our dataset have been annotated and are available in various formats. For object detection, we use the VOC \cite{everingham2010pascal} and COCO \cite{lin2014microsoft} formats; For semantic segmentation, we use the VOC format; For instance segmentation, we use the COCO format. We have made our training and validation sets, alongwith ground-truth annotations, publicly available. To evaluate the performance of algorithms on our test set, researchers can submit their results via our online benchmark suite. This feature serves a dual purposes: it not only streamlines the process of comparing and validating various algorithms but also cultivates a collaborative environment within the research community, promoting the exchange of methods and results. Through its automated and standardized evaluation mechanism, we believe that our benchmark suite represents a significant step forward in the field.

\section{Experimental Results}
\label{sec.experiments}

\subsection{Experimental Setup}
\label{sec.experimental_setup}

\begin{figure*}[t!]
	\begin{center}
		\centering
		\includegraphics[width=1\textwidth]{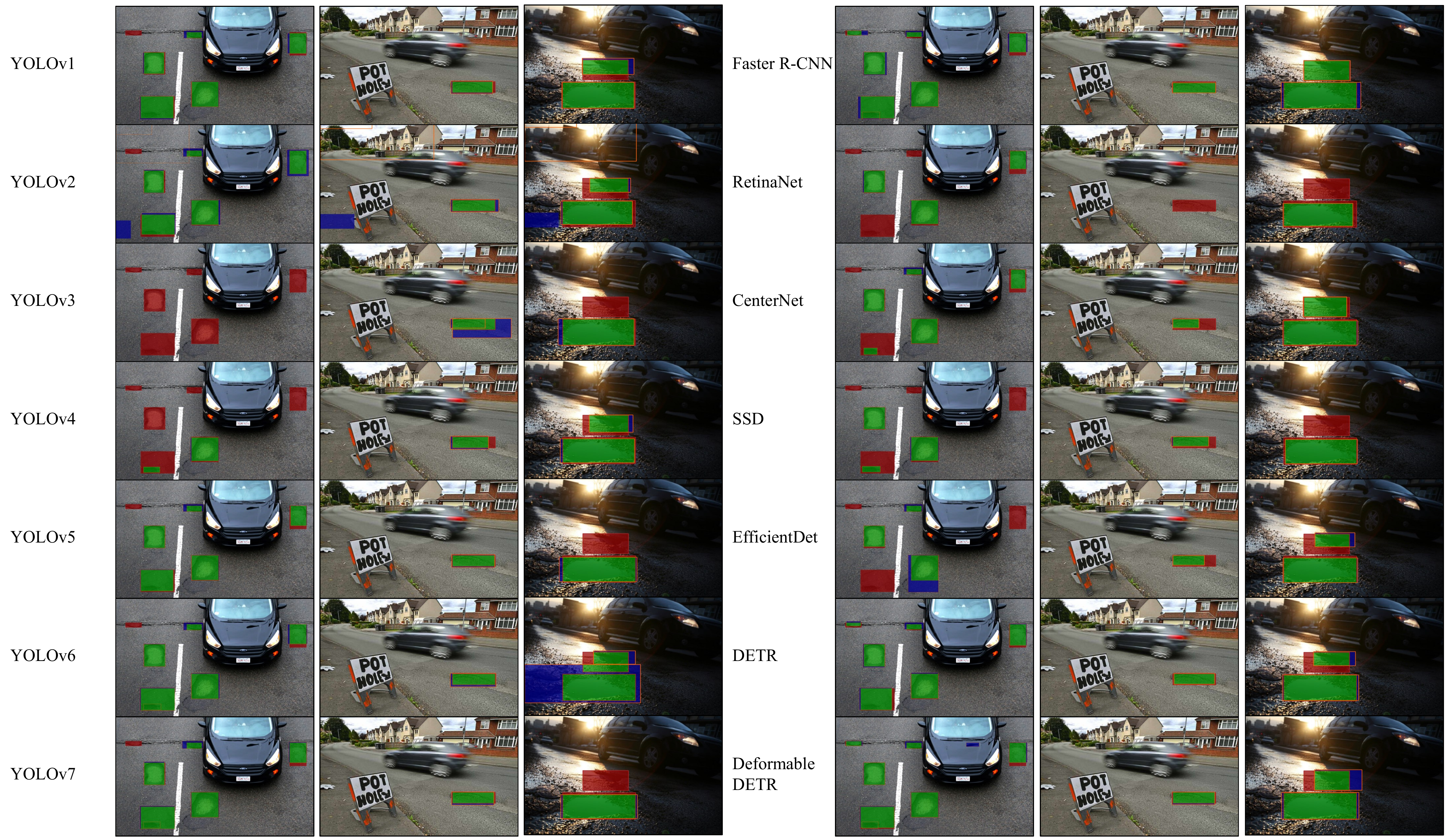}
		\centering
		\caption{Qualitative experimental results of object detection. The green areas in the image represent true-positive predictions, the blue areas represent false-positive predictions, and the red areas represent false-negative predictions.}
		\label{fig.objpic}
	\end{center}
\end{figure*}

\begin{figure*}[t!]
	\begin{center}
		\centering
 		\includegraphics[width=0.99\textwidth, height=1.25\textwidth]{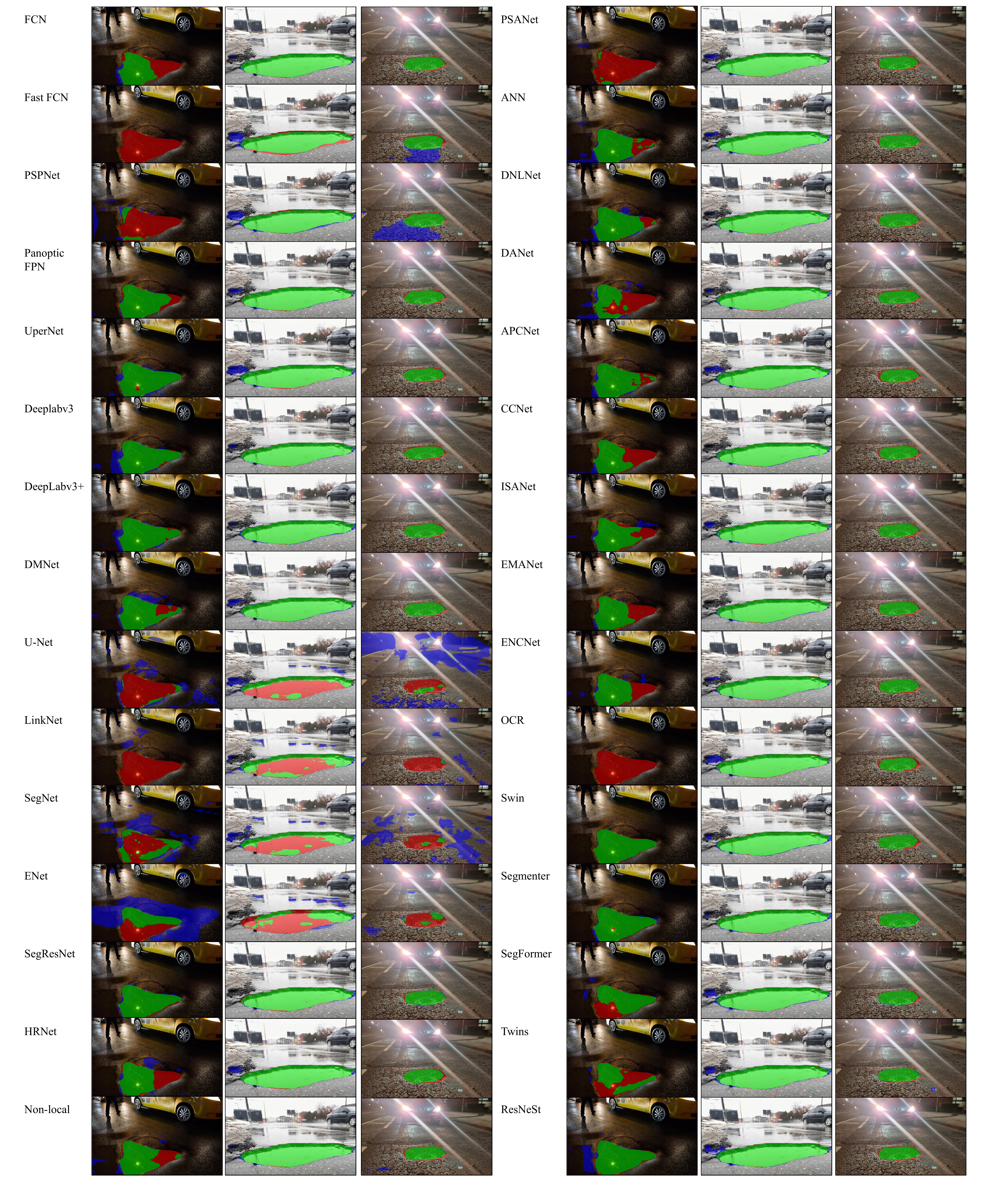}
		\centering
		\caption{Qualitative experimental results of semantic segmentation. The green areas in the image represent true-positive predictions, the blue areas represent false-positive predictions, and the red areas represent false-negative predictions.}
		\label{fig.segpic1}
	\end{center}
\end{figure*}

We \clr{conduct extensive baseline experiments} using 14 object detection networks, 30 semantic segmentation networks, and ten instance segmentation networks. All networks \clr{are} implemented in PyTorch. All experiments \clr{are} conducted on \clr{an} NVIDIA RTX 3090 GPU and an Intel Xeon Platinum 8255C CPU. Each network \clr{is} trained for 150 epochs. We keep the default settings of each network. The quantitative results of object detection, semantic segmentation, and instance segmentation are presented in Tables \ref{table.object_detection_result}, \ref{table.semantic_segmentation_result}, and \ref{table.instance_segmention_result}, \clr{respectively}. Additionally, the qualitative experimental results of object detection, semantic segmentation, and instance segmentation are shown in Figs. \ref{fig.objpic}, \ref{fig.segpic1}, and \ref{fig.instancepic}, \clr{respectively}.

\begin{table*}[]\
		\settablefont
		\centering
		\caption{Quantitative comparison of Object Detection networks, with the best results shown in bold type.}
  {
        \begin{tabular}{l|c|c|ccccc|ccccc}
			\toprule[1.pt]
			\renewcommand{\arraystretch}{2}
			\multirow{2}*{Network} &\multirow{2}*{Params} &\multirow{2}*{FPS} & \multicolumn{5}{c|}{Validation Set} & \multicolumn{5}{c}{Test Set}  \\
			\cline{4-13} 
&&& AP$_S$ (\%) ↑ & AP$_M$ (\%) ↑ & AP$_L$ (\%) ↑ & AP$_{A}$ (\%) ↑ & mIoU (\%) ↑ & AP$_S$ (\%) ↑ & AP$_M$ (\%) ↑ & AP$_L$ (\%) ↑ & AP$_{A}$ (\%) ↑ & mIoU (\%) ↑ \\
			\cline{1-13}
      
YOLOv1 \cite{redmon2016you} &28.49 M&84.36&13.40&32.10&45.10&35.10&54.70&10.70&25.20&45.10&33.60&53.90 \\
YOLOv2 \cite{redmon2017yolo9000} &28.52 M&23.67&14.30&39.90&53.80&42.10&60.50&11.20&28.10&46.10&34.50&57.60 \\
YOLOv3 \cite{redmon2018yolov3} &61.52 M&55.33&24.20&44.80&55.90&47.40&69.70&26.60&43.00&56.00&47.50&66.00 \\
YOLOv4 \cite{bochkovskiy2020yolov4} &63.94 M&36.05&31.10&47.60&66.80&56.60&76.70&31.30&49.00&61.30&52.60&74.30 \\
YOLOv5 \cite{jocher2022ultralytics} &46.14M&36.84&42.70&60.20&74.10&65.30&76.60&40.70&56.50&67.60&59.80&73.60 \\
YOLOv6 \cite{li2022yolov6} &58.47 M&48.73&48.60&\textbf{66.10}&77.10&\textbf{69.60}&83.90&52.20&\textbf{66.60}&\textbf{76.40}&\textbf{69.60}&\textbf{82.90} \\
YOLOv7 \cite{wang2022yolov7} &37.20 M&49.20&45.30&59.70&73.90&65.50&82.70&44.90&57.90&74.90&65.40&78.80 \\
Faster R-CNN \cite{ren2015faster} &136.69 M&25.29&21.40&52.30&64.40&55.00&77.00&21.30&54.00&63.50&53.30&75.30 \\
RetinaNet \cite{lin2017focal} &36.33 M&34.78&25.90&51.40&58.20&50.50&58.00&22.70&40.70&54.90&45.50&56.30 \\
CenterNet \cite{zhou2019objects} &191.24 M&23.38&27.80&53.40&66.30&55.90&75.60&29.50&52.70&64.30&55.10&74.30 \\
SSD \cite{liu2016ssd} &23.61M&136.73&29.10&54.80&69.90&59.30&79.90&30.80&53.90&70.00&59.10&78.10 \\
EfficientDet \cite{DBLP:conf/cvpr/TanPL20} &8.01 M&12.89&41.20&54.50&74.20&63.60&79.60&40.60&57.40&73.30&64.10&79.00 \\
DETR \cite{carion2020end} &41.30 M&30.80&43.20&62.40&\textbf{80.50}&68.90&\textbf{85.90}&45.30&61.00&74.40&65.40&81.30 \\
Deformable DETR \cite{zhu2020deformable} &39.85 M&31.60&\textbf{50.20}&63.50&79.60&69.50&82.60&\textbf{56.40}&61.50&75.00&68.10&80.60 \\
			\bottomrule[1.pt]
			\end{tabular}
			}
                \label{table.object_detection_result}
\end{table*}
 
\subsection{Evaluation Metrics}
\label{sec.evaluation_metrics}
\clr{We use average precision (AP) and mean intersection over union (mIoU) as the evaluation metrics in the object detection task. Furthermore, we utilize accuracy (Acc), IoU, precision (Pre), recall (Rec), and F$1$-score (Fsc) to quantify the performance of semantic segmentation networks. Moreover, we use AP as the evaluation metric in the instance segmentation task.

Following the evaluation on the MS COCO \cite{lin2014microsoft} dataset, we compute AP with respect to different sizes of road potholes: small (denoted as AP$_S$), medium (denoted as AP$_M$), and large (denoted as AP$_L$). We use two proportion thresholds to determine small, medium, and large road potholes. Referring to Fig. \ref{fig.trainvaltest}, the first 300 potholes are considered small (with a pothole area proportion of less than $1.12\%$); The potholes numbered from 301 to 600 (with a pothole area proportion between $1.12\%$ and $3.72\%$) are considered medium; the remaining potholes are considered large (with a pothole area proportion greater than $3.72\%$). Additionally, AP$_A$ represents an average AP score that provides a comprehensive evaluation of the model's performance across all pothole sizes.}

\begin{table*}[t] 
		\settablefont
		\centering
		\caption{Quantitative comparison of semantic segmentation networks, with the best results shown in bold type.}
  { 
        \begin{tabular}{l|c|c|ccccc|ccccc}
			\toprule[1.pt]
			\renewcommand{\arraystretch}{2}
			\multirow{2}*{Network} &\multirow{2}*{Params} &\multirow{2}*{FPS} & \multicolumn{5}{c|}{Validation Set} & \multicolumn{5}{c}{Test Set}  \\
			\cline{4-8} \cline{9-13} 
			&&& IoU (\%) ↑ &  Fsc (\%) ↑ & Pre (\%) ↑ &  Rec (\%) ↑ & Acc (\%) ↑& IoU (\%) ↑ &  Fsc (\%) ↑ & Pre (\%) ↑ &  Rec (\%) ↑ & Acc (\%) ↑  \\
			\cline{1-13}
			
			FCN	\cite{long2015fully} &	49.48 M&16.54&			 79.60&	 88.64&	 91.81&	 85.69&	 85.69&			 74.96&	 85.96&	 87.39&	 84.05&	 97.25\\	
			Fast FCN	\cite{Wu2019}&	87.85 M&10.47	&		 81.28&	 89.67&	 92.33&	 87.17&	 96.93&			 76.63&	 86.77&	 88.57&	 85.04&	 97.47\\
			PSPNet \cite{Zhao2017}&		48.98 M&	16.14&		 81.90&	 90.05&	 92.23&	 87.96&	 97.34&			 74.93&	 85.67&	 85.05&	 86.30&	 97.18\\
			Panoptic FPN   \cite{kirillov2019panoptic}&	 28.51 M& 18.46& 77.14&	 87.10&	 90.91&	 83.59&	 96.61&			 73.23&	 84.55&	 85.14&	 83.69&	 97.07\\
			UperNet \cite{Xiao2018}&	41.4 M&	17.20	&		 79.97&	 88.87&	 89.97&	 87.81&	 96.99&			 73.72&	 84.87&	 84.51&	 85.24&	 96.98\\
			DeepLabv3 \cite{florian2017rethinking}& 68.11 M&13.07& 79.79&	 88.76&	 91.81&	 85.90&	 97.02&		     76.86&	 86.90&	 87.75&	 86.10&	 97.43\\
			DeepLabv3+ \cite{Chen2018}&  	43.58 M&	16.15&		 81.16&	 89.60&	 89.64&	 89.57&	 97.16&			 75.88&	 86.28&	 84.52&	 88.13&	 97.28\\
			DMNet \cite{He2019a}& 	53.28 M&	14.52&	 79.82&	 88.78&	 92.63&	 85.23&	 97.05&			 74.39&	 85.32&	 89.11&	 81.83&	 97.23\\
			U-Net \cite{ronneberger2015u}& 	29.06 M&12.10	&	 69.42&	 81.92&	 83.35&	 80.62&	 94.57&			 64.12&	 78.15&	 78.74&	 78.13&	 94.41\\
			LinkNet	\cite{Chaurasia2017}& 	44.00 M&	11.70&	 76.23&	 86.51&	 86.42&	 86.57&	 96.14&			 72.83&	 84.24&	 84.25&	 84.27&	 95.72\\
			SegNet	\cite{Badrinarayanan2017}&  212.96 M&	1.23& 71.62&	 83.44&	 84.73&	 82.21&	 95.11&			 66.93&	 80.10&	 81.32&	 79.06&	 94.42\\
			ENet \cite{Paszke2016}&  1.36 M&	21.98&		 74.23&	 85.10&	 86.21&	 83.93&	 95.70&			 74.91&	 85.62&	 86.95&	 84.38&	 96.03\\
			SegResNet \cite{Myronenko2018}&	204.31 M&	2.45 &	 81.91&	 90.14&	 90.91&	 82.65&	 97.13&			 79.25&	 88.34&	 88.53&	 88.29&	 96.82\\
			HRNet \cite{Sun2019}&	65.86 M&22.46&		 80.46&	 89.17&	 91.77&	 86.73&	 97.12&			 76.63&	 86.77&	 88.57&	 85.04&	 97.73\\
			Non-local \cite{Wang2018}&	50.03 M&14.13&			 80.82&	 89.40&	 89.07&	 89.72&	 97.09&			 71.99&	 83.72&	 83.12&	 84.32&	 96.74\\
			PSANet \cite{Zhao2018}&	48.98 M&12.57&		 80.67&	 89.30&	 93.10&	 85.80&	 97.19&			 76.43&	 86.64&	 86.91&	 86.37&	 97.36\\
			ANN	\cite{zhu2019asymmetric}&	46.23 M&	15.18 &				 80.91&	 89.44&	 88.50&	90.41&	 97.08&		 76.70&	 86.82&	 87.94&	 85.72&	 97.43\\
			DNLNet	\cite{Yin2020}&	50.13 M&10.15&			 79.36&	 88.49&	 90.95&	 86.16&	 96.93&			 73.26&	 84.57&	 86.75&	 82.49&	 97.01\\
			\cl{DANet}	\cite{Fu2019}&		49.82 M& 14.37 &		 74.47&	 85.37&	 91.03&	 80.37&	 96.23&			 73.19&	 84.52&	 85.75&	 83.33&	 97.01\\
			APCNet	\cite{He2019}&		56.46 M&14.16&		 81.05&	 89.53&	\textbf{94.06}&	 85.42&	 97.27&			 75.23&	 85.87&	 88.66&	 83.24&	 97.27\\
			\cl{CCNet}	\cite{Huang2019}&	49.83M&	15.70&			 80.40&	 89.13&	 92.43&	 86.07&	 97.13&			 75.93&	 86.32&	 89.27&	 83.56&	 97.36\\
			ISANet \cite{huang2000interlaced}&	37.69 M&	18.85&		 80.60&	 89.26&	 91.24&	 87.19&	 97.13&			 76.15&	 86.46&	 88.99&	 84.07&	 97.40\\
			EMANet	\cite{Li2019}&	42.09 M&	18.17	&	 80.71&	 89.32&	 93.16&	 85.79&	 97.19&			 75.02&	 85.73&	\textbf{90.01}&	 81.83&	 97.29\\
			ENCNet	\cite{Zhang2018}&		35.89 M&	17.99&	 79.90&	 88.83&	 91.66&	 86.17&	 97.04&			 74.12&	 85.13&	 84.68&	 85.59&	 97.10\\
			OCR	\cite{yuan2020object}&	12.08 M&27.76&		 63.57&	 77.73&	 81.26&	 74.50&	 94.16&			 62.59&	 76.99&	 76.09&	 77.91&	 95.35\\
			Swin \cite{Liu2021}&		121.3 M& 6.55	&	 82.70&	 90.53&	 90.45&	 90.16&	 97.41&			 77.98&	 87.63&	 85.04&	 90.37&	 97.47\\
			Segmenter \cite{Strudel2021}&	102.5 M&9.52& \textbf{83.21}&	\textbf{90.81}&	 91.26&	\textbf{90.41}&	\textbf{97.50}& \textbf{80.74}	&  \textbf{89.34}&	 87.25&			\textbf{91.54}&	\textbf{97.83}\\
			SegFormer \cite{Xie2021}&	3.75 M&	15.51&	 73.43&	 84.68&	 87.99&	 81.61&	 95.96&			 74.91&	 85.65&	 87.11&	 84.25&	 97.27\\
			Twins \cite{Chu2021}&	47.58 M&	7.96&			 81.57&	 89.85&	 91.32&	 88.42&	 97.27&			 79.02&	 86.38&	 87.94&	 84.87&	 97.36\\
			ResNeSt	\cite{Zhang2022} &	90.91 M& 9.48 &		 79.66&	 88.68&	 91.34&	 86.18&	 96.99&			 79.03&	 88.29&	 88.83&	 87.75&	 97.69\\
			
			\bottomrule[1.pt]
			\end{tabular}}
                \label{table.semantic_segmentation_result}
	\end{table*}

\begin{table*}[t]\
		\settablefont
		\centering
		\caption{Quantitative comparison of instance segmentation networks, with the best results shown in bold type.}
            {
		\begin{tabular}{l|c|c|cccc|cccc}
			\toprule[1.pt]
			\renewcommand{\arraystretch}{2}
			\multirow{2}*{Network} &\multirow{2}*{Params}&\multirow{2}*{FPS} & \multicolumn{4}{c|}{Validation Set} & \multicolumn{4}{c}{Test Set}  \\
			\cline{4-7} \cline{7-11} 
&&& AP$_S$ (\%) ↑ & AP$_M$ (\%) ↑ & AP$_L$ (\%) ↑ & AP$_{A}$ (\%) ↑ & AP$_S$ (\%) ↑ & AP$_M$ (\%) ↑ & AP$_L$ (\%) ↑ & AP$_{A}$ (\%) ↑  \\
\hline
      
Mask R-CNN \cite{he2017mask} &43.97M&31.00 &41.30&29.40& \textbf{61.70} &53.70& \textbf{31.00} & \textbf{55.60} & \textbf{62.40} & \textbf{52.10} \\
DCN \cite{dai2017deformable} &97.30M& 15.20&5.10&25.10&59.30&50.10&5.00&24.20&55.10&46.30 \\
GCNet \cite{cao2019gcnet} &100.00M&12.40&15.60& \textbf{34.10} &60.90&53.60&10.10&30.30&57.20&49.50 \\
YOLACT \cite{bolya2019yolact} &34.73M&2.80&22.10&24.20&53.40&45.70&2.00&17.80&39.10&33.10 \\
Cascade R-CNN \cite{cai2019cascade} &77.02M&24.20 &33.20&33.20&61.20& \textbf{54.10} &12.70&30.60&54.80&47.70 \\
Mask Scoring R-CNN \cite{huang2019mask} &60.23M&30.40 &\textbf{46.60}&32.00&58.80&51.60&12.20&32.80&55.30&48.60 \\
SOLO \cite{wang2020solo} &36.12M&30.40 &0.80&15.60&52.70&42.80&1.70&18.50&45.70&37.80 \\
SOLOv2 \cite{wang2020solov2} &46.23M&31.70 &8.30&20.00&49.80&41.50&2.00&21.10&48.80&40.70 \\
CondInst \cite{tian2020conditional} &34.16M&26.00&31.70&27.40&58.20&49.80&4.50&26.70&53.30&45.00 \\
BoxInst \cite{tian2021boxinst} &34.96M& 21.70&8.90&22.60&52.80&44.40&6.10&22.50&48.20&40.60 \\

			\bottomrule[1.pt]
			\end{tabular}}

                \label{table.instance_segmention_result}
	\end{table*}

\subsection{Object Detection Network Performance}
\label{sec.Object Detection Network Performance}

\clr{
The quantitative results of object detection networks presented in Table \ref{table.object_detection_result} suggest the following findings: (1) YOLOv6 achieves the highest AP$_A$ and AP$_M$, and Deformable DETR achieves the highest AP$_S$ on both the validation and test sets; (2) regarding AP$_L$ and mIoU, DETR outperforms others on the validation set, while YOLOv6 leads with the highest scores on the test set; (3) YOLOv6 demonstrates real-time performance, slightly faster than DETR and Deformable DETR. 

Although most networks are capable of detecting larger potholes, they often struggle with medium and smaller ones. Transformer-based networks, such as DETR and Deformable DETR, consistently outperform CNN-based networks in detecting small potholes. This superiority can be attributed to the fact that small potholes might become undetectable after passing through multiple convolution layers in CNN-based models. In contrast, Transformer-based approaches, with a self-attention mechanism, can effectively capture information from all positions without relying on pooling operations, allowing them to preserve important details relevant to small potholes. However, the generalizability of Transformer-based networks is limited, likely due to data availability constraints. While these networks show promise for small object detection, further research and data collection efforts may be necessary to enhance their performance across diverse scenarios.

It is also worth noting that YOLOv6 demonstrates a higher occurrence of false-positive regions during nighttime, as illustrated in Fig. \ref{fig.objpic}. This indicates its potential limitations when dealing with extreme or low-light situations. However, we still recommend YOLOv6 as the preferred choice for object detection-based road pothole detection, primarily due to its advantageous balance between speed and accuracy.
}

\subsection{Semantic Segmentation Network Performance}
\label{sec.Semantic Segmentation Network Performance}

\begin{figure*}[t!]
	\begin{center}
		\centering
		\includegraphics[width=0.99\textwidth]{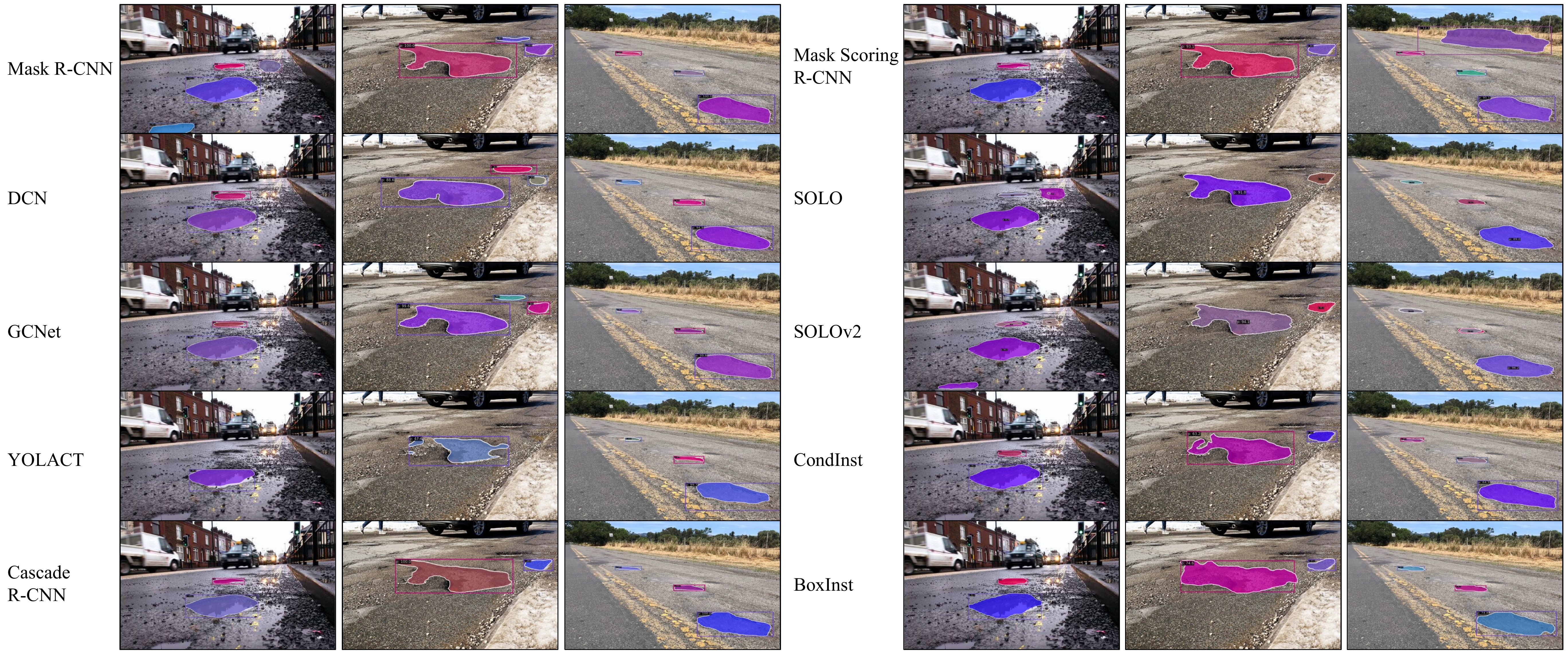}
		\centering
		\caption{Qualitative experimental results of instance segmentation. Different road potholes are shown in different colors.}
		\label{fig.instancepic}
	\end{center}
\end{figure*}

\clr{
As shown in Table \ref{table.semantic_segmentation_result}, Segmenter consistently achieves the highest IoU, Fsc, Rec, and Acc on both the validation and test sets. The results presented in Fig. \ref{fig.segpic1} further illustrate that Segmenter can yield the most accurate boundaries, with minimal occurrences of false-positive regions, even under poor illumination conditions. Additionally, APCNet and EMANet stand out by achieving the highest Pre on the validation and test sets, respectively. Regrettably, these networks all fall short in achieving real-time performance, which imposes limitations on their practical applicability in real-world scenarios. 

This article represents a pioneering effort in utilizing Transformer-based networks for road pothole detection. Except for SegFormer, Transformer-based networks demonstrate superior performance compared to the CNN-based approaches. This superiority can be attributed to two key factors. First, Transformers are inherently designed to efficiently capture long-range dependencies and global context. In semantic segmentation, understanding the relationships between distant pixels or objects holds significant importance. Transformers are adept at modeling these global relationships, allowing for more context-aware segmentation. Furthermore, Transformers rely on attention mechanisms that can capture fine-grained spatial relationships within an image. This capability can lead to more precise and context-aware segmentation, particularly when dealing with densely packed objects or those with intricate shapes.

We also compare the generalizability of these networks. DeepLabv3, ENet, SegResNet, DANet, OCR, Segmenter, SegFormer, Twins, ResNeSt demonstrate comparable performance on both the validation and test sets, with the IoU fluctuating within a range of $0.06\%$ to $2.93\%$. Among them, ENet achieves the best generalizability on our UDTIRI benchmark.  
}

\subsection{Instance Segmentation Network Performance}

\clr{
The results presented in Table \ref{table.instance_segmention_result} indicate Mask Scoring R-CNN, GCNet, Mask R-CNN, and Cascade R-CNN achieve the highest AP$_S$, AP$_M$, AP$_L$, and AP$_A$, respectively, on the validation set, while Mask R-CNN outperforms all other models across all evaluation metrics on the test set. These findings suggest the superior segmentation accuracy of two-stage models when compared to one-stage models. Additionally, as illustrated in Fig. \ref{fig.instancepic}, Mask R-CNN consistently produces accurate road pothole boundaries, even under challenging conditions characterized by high humidity and strong light reflections. Furthermore, it is noteworthy that Mask R-CNN achieves comparable efficiency to one-stage approaches, such as SOLO and SOLOv2, making it a practical choice, especially for resource-limited hardware. Nonetheless, it is evident that all the instance segmentation networks compared in this study achieve unsatisfactory performance, particularly when segmenting small road potholes, underscoring the critical need for further research and improvement in this area.
}

\section{Discussion}
\label{sec.discussion}
\clr{
This study has two notable limitations. First, the current benchmark primarily focuses on single-model road pothole detection, without exploring the potential benefits of multi-sensor data fusion. Future iterations of our benchmark will incorporate additional spatial geometric information and comprehensively investigate data-fusion networks, providing a more comprehensive evaluation of model performance. Secondly, our benchmark currently focuses exclusively on potholes, omitting the inclusion of other common road damages, such as cracks. Detecting cracks is essential not only for urban road maintenance but also for automated driving perception systems. To enhance the comprehensiveness of our benchmark and align it more closely with real-world scenarios, it is crucial to incorporate additional datasets comprising various road damage types and evaluate a wider range of models for crack detection.
}

\section{Conclusion}
\label{sec.Conclusion}

In this article, we introduced an online open-source benchmark suite, referred to as UDTIRI, within which the first intelligent road inspection competition -- road pothole detection was launched. The competition provides a large-scale, well-annotated dataset that can be used for the training and evaluation of object detection, semantic segmentation, and instance segmentation networks. The annotations for the training and validation sets are made accessible to researchers, where a comprehensive performance evaluation of their developed networks on the test set can be obtained by submitting the results through our online benchmark platform. Furthermore, we provided extensive baseline experimental results using 14 object detection networks, 30 semantic segmentation networks, and ten instance segmentation networks. With upcoming IRI competitions set to be introduced within the UDTIRI benchmark, we believe that our benchmark will act as a driving force, encouraging the integration of advanced UDT techniques into IRI.

\normalem
\bibliographystyle{IEEEtran}      

\end{document}

%% file: packages.tex
\usepackage{ulem}
\usepackage{url}

\usepackage{scalerel}
\usepackage{color}
\usepackage{tikz}
\usetikzlibrary{svg.path}
\definecolor{orcidlogocol}{HTML}{A6CE39}
\tikzset{
  orcidlogo/.pic={
    \fill[orcidlogocol] svg{M256,128c0,70.7-57.3,128-128,128C57.3,256,0,198.7,0,128C0,57.3,57.3,0,128,0C198.7,0,256,57.3,256,128z};
    \fill[white] svg{M86.3,186.2H70.9V79.1h15.4v48.4V186.2z}
                 svg{M108.9,79.1h41.6c39.6,0,57,28.3,57,53.6c0,27.5-21.5,53.6-56.8,53.6h-41.8V79.1z M124.3,172.4h24.5c34.9,0,42.9-26.5,42.9-39.7c0-21.5-13.7-39.7-43.7-39.7h-23.7V172.4z}
                 svg{M88.7,56.8c0,5.5-4.5,10.1-10.1,10.1c-5.6,0-10.1-4.6-10.1-10.1c0-5.6,4.5-10.1,10.1-10.1C84.2,46.7,88.7,51.3,88.7,56.8z};
  }
}
\newcommand\orcidicon[1]{\href{https://orcid.org/#1}{\mbox{\scalerel*{
\begin{tikzpicture}[yscale=-1,transform shape]
\pic{orcidlogo};
\end{tikzpicture}
}{|}}}}
\usepackage{hyperref}
\hypersetup{
    colorlinks=true,
    linkcolor=blue,
    filecolor=black,      
    urlcolor=magenta,
    citecolor=purple
}
\usepackage{tikz-network}